\documentclass[preprint, letterpaper]{IEEEtran}
\usepackage{amsmath,amsfonts}
\usepackage{algorithmic}
\usepackage{array}
\usepackage[caption=false,font=normalsize,labelfont=sf,textfont=sf]{subfig}
\usepackage{textcomp}
\usepackage{stfloats}
\usepackage{amssymb}
\usepackage{url}
\usepackage{verbatim}
\usepackage{graphicx}
\usepackage{mathrsfs} 
\usepackage{booktabs}
\usepackage{multirow}
\usepackage{makecell} 
\usepackage{array}
\usepackage{cite}

\def\BibTeX{{\rm B\kern-.05em{\sc i\kern-.025em b}\kern-.08em
    T\kern-.1667em\lower.7ex\hbox{E}\kern-.125emX}}
\usepackage{balance}
\begin{document}
\title{ATN3D: Density-Aware LiDAR-Radar Early 3D Object Detection Under Extreme Sparsity} 
\author{Debojyoti~Biswas~\IEEEmembership{Member,~IEEE}, and Xianbiao~Hu}

\markboth{}%
{ATN3D: Density-Aware LiDAR-Radar Early 3D Object Detection Under Extreme Sparsity}
\maketitle

\begin{abstract}
3D object detection is the backbone of perception for automated vehicles (AV) and broader intelligent transportation systems applications. Long-range detection is challenging because sensing evidence is sparse; yet this ``long-range'' scenario is routine in traffic. Although \(>\!30\,\mathrm{m}\) is often labeled long-range in computer vision, on roadways it affords only \(\approx 1\!-\!2\,\mathrm{s}\) for perception and decision-making. Under such extreme sparsity, two core challenges arise. First, early multimodal fusion tends to discard sparsity information and inject noise from empty or falsely occupied cells, degrading long-range recall. Second, context-agnostic uniform channel supervision favors dense and near-range samples, leaving far and small objects under-optimized, delaying the earliest detection of distant objects. We propose ``Ask The Neighbor'' (ATN3D), a LiDAR-Radar framework tailored for sparse-range conditions. ATN3D introduces (i) Density-aware early fusion with cross-modal gating that conditions fusion on per-voxel density/sparsity and Radar evidence, (ii) Occupancy-gated neighborhood aggregation with circular kernels to aggregate only from credible cells, (iii)  Evidence-conditioned channel self-attention to adapt channel weights with weather/range, and (iv) a Range-aware loss that re-balances classification and localization by distance, aligning training with distance-stratified evaluation. On the VoD benchmark across clear and foggy conditions, ATN3D surpasses strong baselines: \(+3.55\%\) mAP in clear weather and \(+8.41\%\) mAP under simulated heavy fog; for \(>\!30\,\mathrm{m}\) objects, gains are \(+3.33\%\) (clear) and \(+2.09\%\) (heavy fog). These results indicate earlier and more reliable long-range detections under sparse sensing in on-road traffic.
\end{abstract}


\section{Introduction}
\label{sec:Intro}
3D object detection serves as the backbone of perception for automated vehicles (AV) and broader intelligent transportation system (ITS) applications: it determines what an AV can safely perceive and thus constrains planning, control, and overall safety margins \cite{hu2023planning}. Reliable 3D detections support downstream tasks, including trajectory prediction, multi-object tracking, and ego-motion planning, across diverse operational design domains (ODDs) \cite{chae2024towards, zhong2022short,guo20223d,sun2025sparsedrive}. To remain accurate, efficient, and robust under varied weather and geography, modern AV stacks fuse complementary sensors (cameras, LiDAR, and 4D Radar) into a consistent scene representation \cite{qian2021robust,ning2024novel, ma2024cam4docc}. Among these, LiDAR’s precise depth and long-range resolution under clear conditions make LiDAR-based 3D detection a production-grade cornerstone \cite{deng2024cmd}.

\begin{figure}[!ht]
\centering
    \includegraphics[width=.47\textwidth]{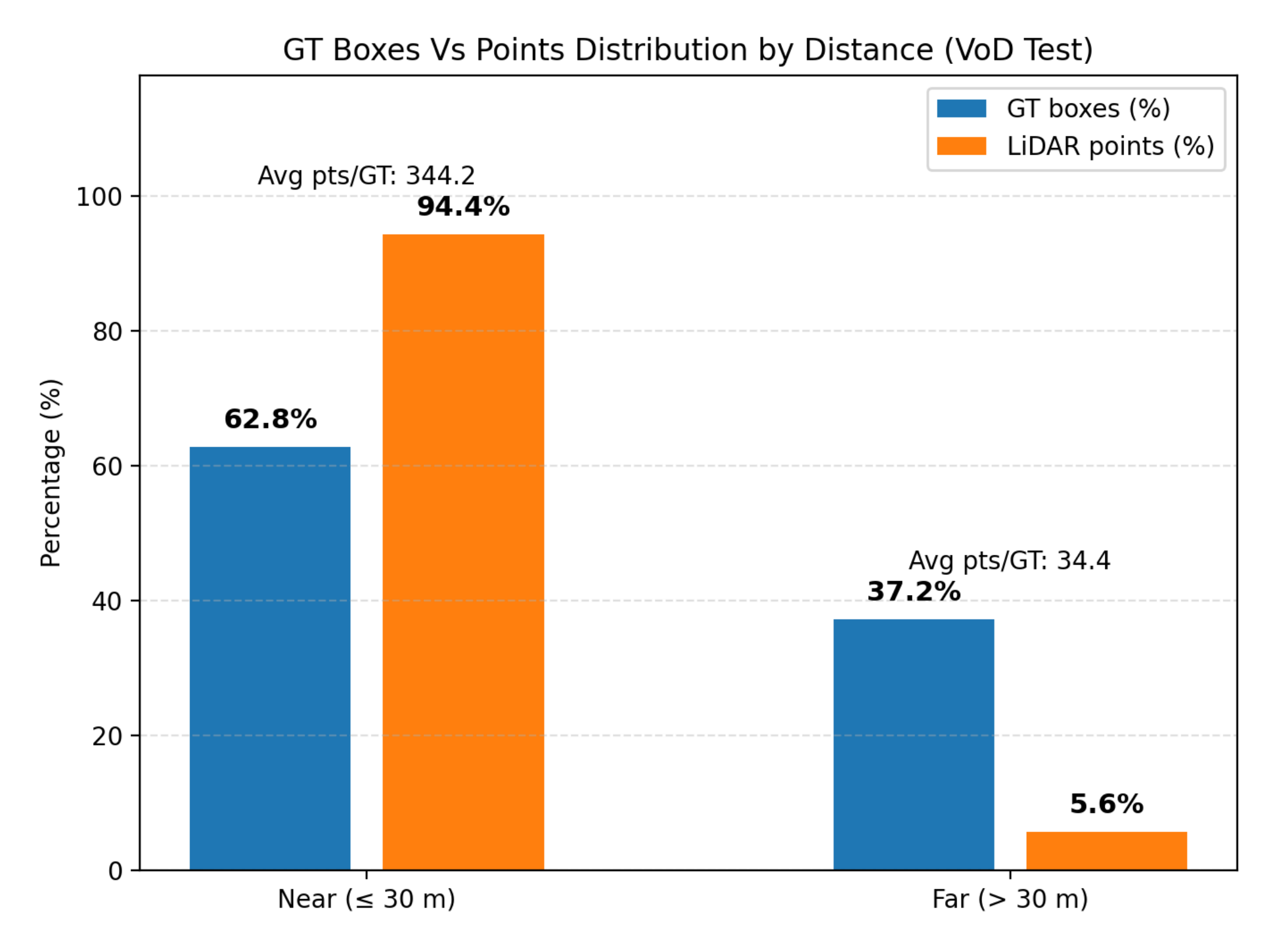}
    \caption{Number of Ground Truth (GT) objects Vs. number of points in near ($\leq$30m) and far ($>$30m) distance in the VoD test set.} 
    \label{fig:gt_radius_plot}
\end{figure}

Long-range 3D detection is challenging due to data sparsity; however, this “long-range” scenario is part of everyday driving for an AV. As shown in Figure \ref{fig:gt_radius_plot} (VoD dataset \cite{apalffy2022}), approximately $37\%$ of objects lie beyond $30\,\mathrm{m}$ yet receive only $\sim6\%$ of LiDAR points; on a per-object basis this yields $344.2$ points at $\leq30\,\mathrm{m}$ versus $34.4$ at $>30\,\mathrm{m}$ ($\approx10\times$ fewer). Unfortunately, $30\,\mathrm{m}$ is not far in traffic: at $40\,\mathrm{mph}$ on an arterial road it corresponds to approximately $1.7\,\mathrm{s}$, and at $75\,\mathrm{mph}$ on a freeway to roughly $0.9\,\mathrm{s}$ of time to react. Thus, while $>30\,\mathrm{m}$ is often labeled long-range in computer vision, on roadways it affords only $1$--$2\,\mathrm{s}$ for perception and decision-making. This mismatch underscores why sparse-range detection is fundamental to AV safety margins.

\begin{figure}[!ht]
\centering
    \includegraphics[width=.47\textwidth]{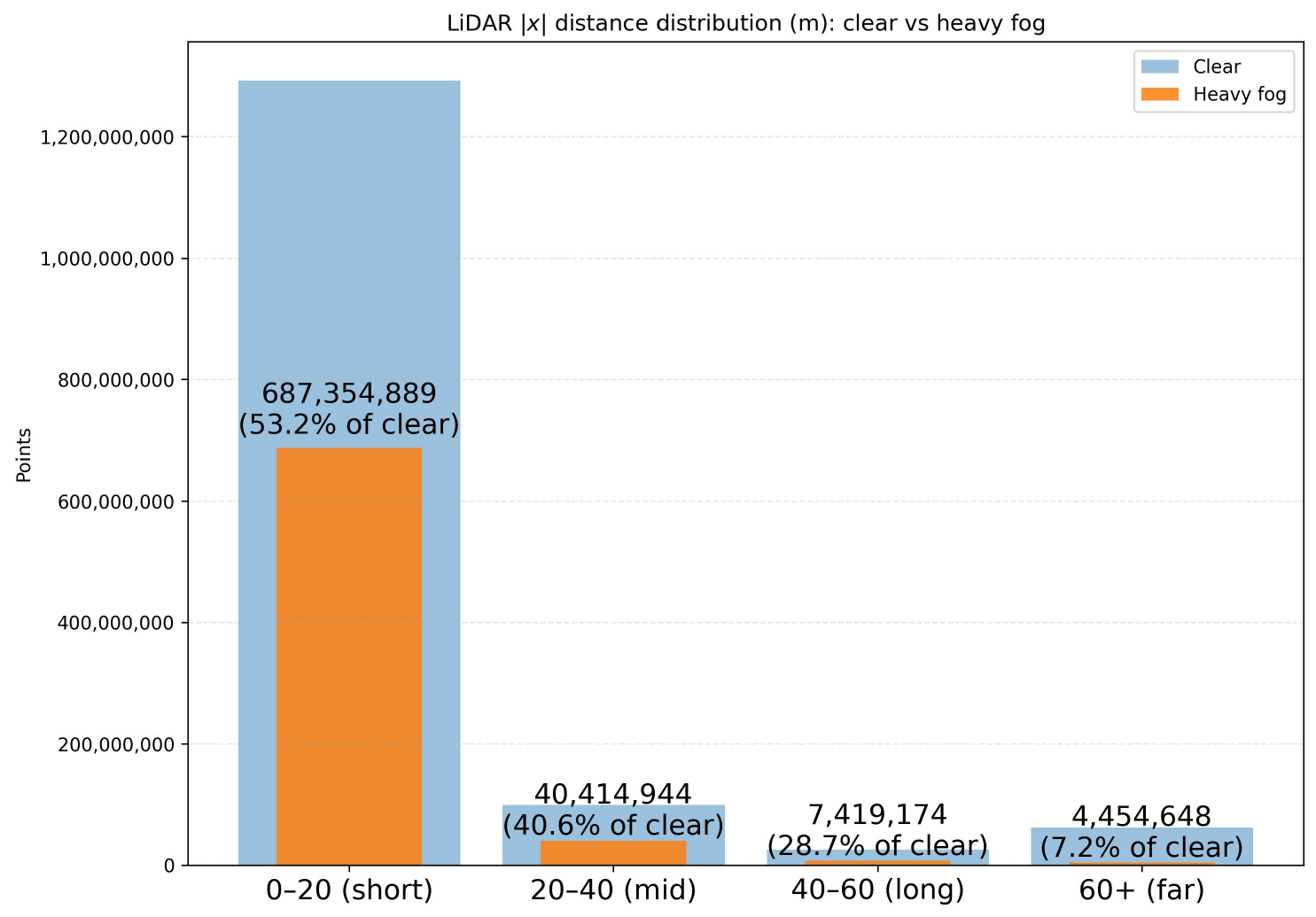}
    \caption{LiDAR point clouds distribution w.r.t different ranges in the VoD clear weather vs. heavy fog weather condition.} 
    \label{fig:sparsity_plot}
\end{figure}

\begin{figure}[!ht]
\centering
    \includegraphics[width=.48\textwidth]{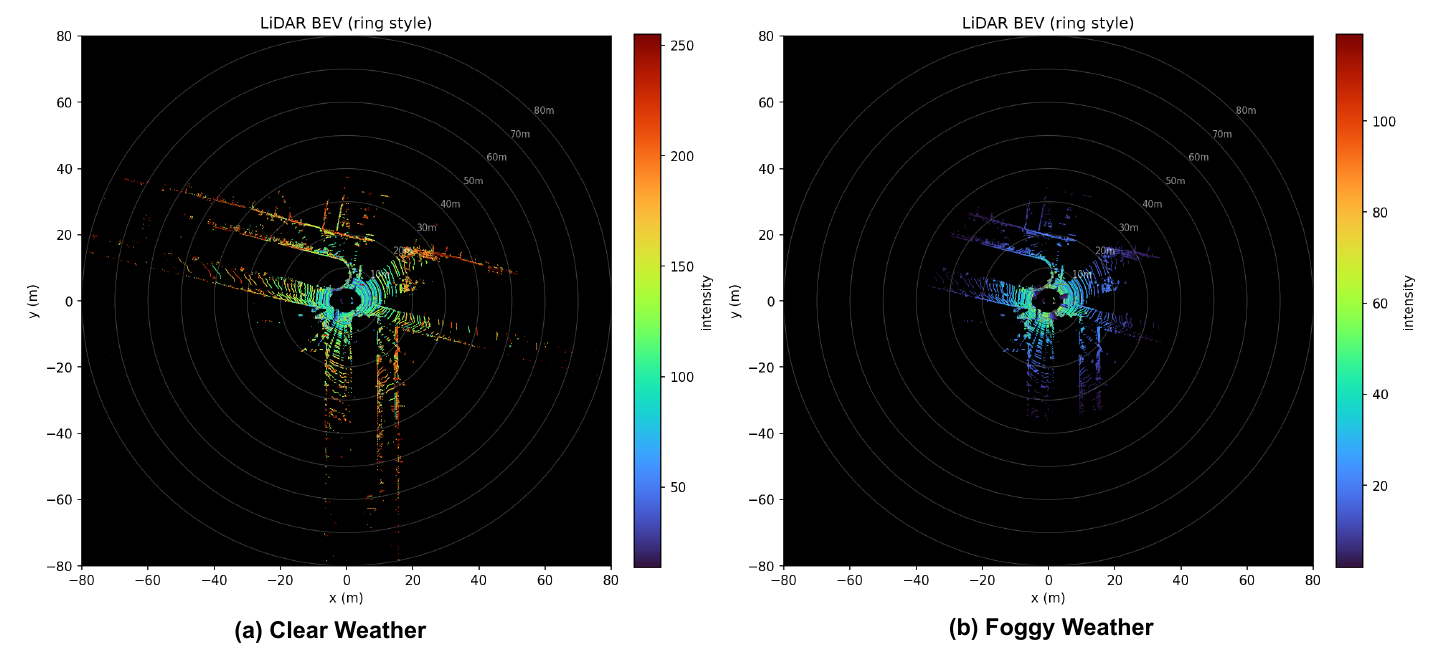}
    \caption{Illustration of LiDAR point clouds in Bird-Eye-View (BEV) for clear vs foggy conditions. The intensity and point range significantly drop in foggy conditions compared to a clear weather LiDAR scene.}
    \label{fig:lidar_plot_clr_vs_fog}
\end{figure}

Worsening the situation, adverse weather further amplifies data sparsity. In practice, this is one of the main reasons why AV deployments often prioritize regions with more benign weather (e.g., milder precipitation and fog), such as Texas and Arizona, rather than northern states like Pennsylvania. As shown in Figure~\ref{fig:sparsity_plot}, heavy fog reduces LiDAR returns by roughly half, making far-range evidence even scarcer; Figure \ref{fig:lidar_plot_clr_vs_fog} likewise shows that both LiDAR intensity and effective range drop markedly in fog compared to clear scenes. While LiDAR–camera fusion is becoming increasingly popular and can moderately improve performance in complex yet clear conditions~\cite{chen2017multi, li2024gafusion}, it does not fully address fog and rain. Prior studies report unsatisfactory performance under heavy fog or rain~\cite{wang2022performance}, and cameras themselves suffer from blur and contrast loss in such conditions, degrading 3D detection and leading to missed detections that threaten AV safety and reliability~\cite{huang2024sunshine, jiang2024weather}.
 
In this study, we advocate LiDAR–Radar perception with an emphasis on early, density-aware fusion and occupancy-preserving aggregation to cope with sparse sensing and adverse-weather degradation. Radar offers weather robustness and Doppler cues \cite{han20234d}, while LiDAR provides precise geometry and long-range resolution, together providing robust sensory data for improved performance. In this study, \emph{sparse-range} refers to operating regimes where per-cell or per-object sensing evidence is scarce (e.g., $10\times$ fewer whether caused by long range or adverse weather). Unless otherwise noted, we treat \emph{extreme sparsity} and \emph{sparse-range} as equivalent descriptions of this condition. Unfortunately, fusion under extreme sparsity is nontrivial. Late BEV-stage fusion can miss fine-grained cross-modal evidence \cite{song2024lirafusion}, and naively stacking Radar with LiDAR does not resolve empty/sparse voxels or cross-modal occupancy mismatch. Specifically, the following two main challenges arise and motivate our design.  

The first core challenge arises from voxel sparsity and the propagation of noise during context aggregation under extreme weather conditions. In the $>30\,\mathrm{m}$ regime and in heavy fog, BEV cells/voxels are predominantly empty or falsely occupied: LiDAR returns collapse, Radar hits are sparse and uneven \cite{wang2022sparse2dense}. Moreover, cross-modal occupancy frequently disagrees due to SNR fluctuations, clutter/multipath, and calibration tolerances. Hence, density-agnostic early fusion results in inferior performance due to incomplete point distributions at the voxel level \cite{wu2025scda}. Also, naive early fusion (e.g., channel concatenation or uniform pooling of Radar into LiDAR features) injects spurious noise by activating non-occupied cells, amplifying sidelobe/ghost responses in LiDAR-Radar empty tiles, and modality dropout that destabilizes optimization (feature statistics shift between different data characteristics and noise distributions) \cite{wang2022multi}. Downstream, vanilla context aggregation (e.g., fixed-shape BEV convolutions or mean/attention pooling) further propagates this contamination, leading to activations leaking from unreliable cells, overly smoothed boundary gradients, and background responses dominating in thin structures (e.g., pedestrians/cyclists) \cite{liu2023you}. The combined effect elevates false-positive rates, delayed localization lock-on at long range, and highly variant gradients in sparse bands, which together depress distance-stratified recall despite seemingly strong aggregate AP.

The second core challenge comes from the uniform supervision across channels in the final BEV dense feature. Adverse weather (e.g., fog) and increasing distance push many samples into sparse-range: LiDAR returns thin out, intensities drop, and per-object evidence collapses, while the relative reliability of Radar cues (Doppler/RCS/SNR) persists. The existing late-fusion strategies, such as Inter-fusion and Intra-fusion methods, apply gating to reduce redundancy in the dense BEV feature; however, they do not control the influence of noise from the background scene, cross-modal domain shifts, or tile-mismatch \cite{huang2025l4dr, song2024lirafusion}. The detector typically applies context-agnostic mixing and directly passes features to the detection head, so channel weights and feature priors remain untuned to the downstream task and drift under heavy fog. On the supervision side, uniform losses and sampling allocate most of the gradient to large, high-evidence boxes; anchor/assignment heuristics select few positives for tiny, distant boxes; thresholds/NMS further suppress low-confidence true-positive predictions \cite{gupta2023far3det}. The combined consequence leads to missed objects, delayed lock-on, and degraded distance-stratified recall and earliest predictions.

To address these challenges, we propose ``Ask The Neighbor'' (ATN3D), a LiDAR–Radar 3D detection framework designed for sparse-range conditions. ATN3D couples evidence-aware early fusion with occupancy-preserving aggregation to prevent noise injection and diffusion under extreme sparsity, and then uses context-adaptive channel weighting and a range-aware objective to counter the systemic bias against long-range targets.

To address the first challenge associated with density-agnostic learning and noise propagation, we introduce two front-end modules. DA-fusion (Density-Aware Early LiDAR–Radar Fusion with Cross-Modal Gating), which augments Pillar Voxel with per-modality density/sparsity and Radar evidence cues (e.g., availability/SNR/Doppler consistency) so that early fusion admits information when local support exists and down-weights or bypasses it when it does not. Also, it provides a comprehensive view of local point distribution statistics, enabling a more focused optimization of weak voxels. O-GNA (Occupancy-Gated Neighborhood Aggregation with Circular Kernels) then performs proximity-weighted context aggregation only from credible, occupied cells. The occupancy gating prevents spurious activations from leaking across receptive fields and keeps object boundaries intact at long range.  

To address the second challenge associated with the evidence-agnostic BEV late-fusion and uniform supervision across the range, we add two back-end modules. E-CSA (Evidence-Conditioned Channel Self-Attention) reweights fused deep channels conditioned on evidence quality and context, suppressing background/clutter channels in heavy fog weather and emphasizing informative cues at long range. RALC (Range-Aware Loss Calculation) jointly rebalances classification and localization losses by distance, aligning optimization with distance-stratified recall and earliest detection, thereby translating gains into earlier, more reliable detections in sparse-range traffic scenes.


The rest of this article is organized as follows. Section~\ref{sec:related} summarizes related work, and Section~\ref{sec:method} introduces the proposed ATN3D framework, with detailed description of the four core technical modules. In Section~\ref{sec:exp}, we introduce the experimental dataset and the proposed ATN3D model is evaluated using the latest SOTA 3D object detection methods on VoD datasets over different weather settings. Finally, Section~\ref{sec:con} summarizes the quantitative findings and outlines future work.

\section{Related Work}
\label{sec:related}

\subsection{LiDAR-only 3D Object Detection}
In recent years, LiDAR has gained popularity as one of the mainstream modalities for 3D object detection on autonomous systems. The growing popularity is due to its long-range receptive field and accurate object dimension information. The detection methods are broadly two kinds: 1) Point-based and 2) Voxel-based. The point-based methods are built upon the PointNet \cite{qi2017pointnet, qi2017pointnet++} series of backbone networks for LiDAR feature extraction. The PointRCNN \cite{shi2019pointrcnn} was a two-stage point-based network. The first stage classifies the foreground and background objects. Next, the second stage processes the region proposals into standard coordinates for GT matching and predictions. Yang et al. \cite{yang20203dssd} proposed 3DSSD, a single-stage point-based 3D detector. The authors improve the performance by introducing the farthest point sampling method. Also, a delicate box prediction network and an anchor-free regression head with a 3D center-ness assignment strategy are proposed for a faster inference rate. However, it was observed that a point-based network is less effective compared to voxel-based and pillar-based networks \cite{palmer2023reviewing}.

The voxel-based methods first convert the irregular point clouds into a regular voxel grid. Each of the voxel are accumulated with several cloud points. Next, 3D CNNs are used to extract a voxel feature map, which is later utilized for 3D object detection. VoxelNet \cite{zhou2018voxelnet} is one of the earliest methods to encode point clouds into voxels. However, the performance was not satisfactory for scenes with large voxel quantities. Yan et al. \cite{yan2018second} introduced sparse convolution to non-empty voxels and significantly reduced the computational cost. Lang et al.  \cite{lang2019pointpillars} proposes a pillar-shaped voxel feature map where the z-axis is collapsed and forms a 2D BEV feature map. On the other hand, Yin et al. \cite{yin2021center} proposed an anchor-free one-stage detector, simplifying the previous 3D detection pipelines. Although the LiDAR-based methods achieved competitive performance, they still face a trade-off between accuracy and efficiency. Also, the performance of LiDAR-based methods degrades in adverse weather conditions, which is crucial for safe autonomous trajectory planning.

\subsection{Radar-only 3D Object Detection}
4D Radar data provides a few additional information, such as range, azimuth, relative velocity, and elevation, making it a much stronger choice for 3D perception in adverse weather conditions. Also, as Radar sensors are less affected by weather conditions, more work started using Radar as the primary modality for 3D object detection and segmentation tasks \cite{danzer20192d, nabati2019rrpn, schumann2018semantic}. For example, Brodeski et al. \cite{brodeski2019deep} use only Radar calibration and propose a new augmentation technique to overcome the limitations of labeled data. Recent advancements, such as 4DRadDet \cite{weng20254draddet}, leverage the raw Radar data and utilize the density information of 4D Radar point clouds to address challenges like data sparsity and noise. The Range and azimuth information have also been utilized in different ways to improve object detection and classification performance \cite{major2019vehicle, decourt2022darod, decourt2024recurrent}. Decourt et al. \cite{9827281} uses the range-Doppler spectra with the Faster R-CNN object detector pipeline for better 3D detection on autonomous vehicles. Wei at al. \cite{wei2022area} pivots on improved temporal/spatial resolutions of mmWave Radars to minimize the false-alarm rate and achieve better accuracy for 3D detection. The temporal relation is also explored by Li et al. \cite{li2022exploiting} to overcome the low angular resolution and precision in recognizing surrounding objects with Radar data. The authors leverage the consistent object properties, such as size, orientation, etc., and propose a temporal relational layer to model the object relationship within successive Radar scans explicitly. Despite performance improvements using solely Radar sensor data, the mentioned methods did not fully utilize the potential information from other modalities. In this study, we focus on exploring and utilizing information from LiDAR and 4D Radar for robust 3D perception in AVs.

\subsection{LiDAR-Radar 3D Object Detection}
Several recent works explored the combination of LiDAR and Radar fusion to leverage features from both modalities toward robust 3D perception for autonomous systems. For example, RadarNet \cite{yang2020Radarnet} used a voxel network for feature extraction for the 3D Radar and LiDAR. This approach misses the elevation information from the 3D Radar sensor and ignores the sparsity in 3D Radar data. Unlike 3D Radar, the 4D Radar provides the spatial and vertical information from the sensor, and is more suitable for accurate 3D object detection \cite{Wang_2023_CVPR}. This version of Radar data gained more attention as it is easier to fuse 3D LiDAR and 4D Radar, with both having the x,y, and z information for each point cloud.

Although the 4D Radar sensors are still developing, there have been a few works in this direction. Interaction-based multi-scale fusion for LiDAR-Radar data is proposed in $M^2$Fusion \cite{wang2022multi}. Additionally, they proposed a data preprocessing method that utilizes a Gaussian distribution to effectively reduce noise in 4D Radar data. Song et al. \cite{song2024lirafusion} proposed to perform early fusion with a joint encoder and middle fusion with an adaptive gated network. A latest method, L4DR \cite{huang2025l4dr}, performs foreground denoising on Radar data to reduce noise from Radar data and introduces a novel Multi-modal encoding technique for bi-directional early fusion on LiDAR-Radar. Yang et al. \cite{yang2024ralibev} introduce RaLiBEV, which proposes a transformer-based enhanced feature fusion and an advanced label assignment strategy for more consistent regression. Moreover, the existing LiDAR-Radar 3D detector uniformly handles the complete scene regardless of weather conditions and point cloud range.

\begin{figure*}[!ht]
\centering
    \includegraphics[width=.95\textwidth]{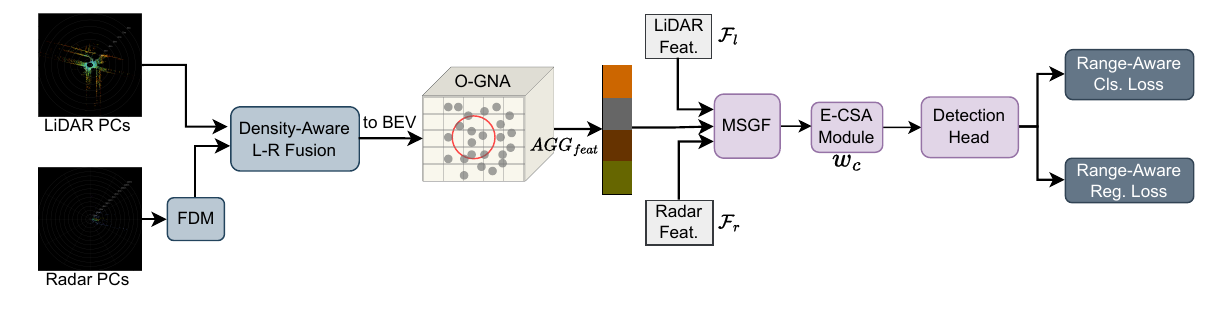}
    \caption{Proposed ATN3D Architecture. Here, FDM = Foreground Denoising Module, O-GNA = Occupancy-Gated Neighborhood Aggregation, E-CSA = Evidence-Conditioned Channel Self-Attention Module, and MSGF= Multi-Scale Gated Fusion.} 
    \label{fig:atn3d_arch}
\end{figure*}

\subsection{3D Object Detection in Sparse-range Scene}
Very few works mention the long-range performance issue in 3D object detection for autonomous vehicles. Far3Det \cite{gupta2023far3det} focuses on developing a well-annotated far-field validation set and proposes an evaluation protocol for 3D object detection. However, obtaining a well-annotated dataset with objects at distances $>60m$ is often difficult and labourous. Khoche et al. \cite{khoche2024towards} address the label imbalance for faraway objects with two 3D detection networks, where one detector focuses on mid-range objects, and another one on long-range objects. This is a LiDAR-only method and not scalable across different modalities. Also, these approaches are not generalized across datasets and do not explicitly tackle the far-field sparsity issue. Wang et al. \cite{wang2022sparse2dense} build a framework that first trains a dense point 3D detector with a dense point cloud as input and designs a sparse point 3D detector with a regular point cloud as input. This is a two-stage training for densifying the point clouds, relying on dense input for the first stage. The LiDAR+Radar training does not provide dense input and is not suitable for such a two-stage training framework. Zhang et al. \cite{zhang2025dsrc} introduce a density-insensitive collaborative multi-agent approach for driving scenes. In this work, the authors construct multi-view dense objects, using ground-truth bounding boxes, to effectively learn a density-insensitive collaborative representation. However, the complete hypothesis relies on the existence of multiple agents in the driving scene. This intuition is often impractical, and inter-agent communication (e.g., cross-talk between multiple sensors) incurs delays and noise, thereby hampering the goal of accurate early object detection. 

To summarize, although recent advances in multimodal fusion and 3D detection have improved robustness under nominal conditions, existing studies have not explicitly addressed voxel-level sparsity under adverse weather or the scarcity of LiDAR returns in far-range voxels from the ego vehicle. Also, despite achieving satisfactory performance in a particular dataset, the frameworks are not scalable across datasets and are not suitable for real-time far object detection. Finally, most frameworks treat the entire scene uniformly, without differentiating voxel quality or sensing range. Effectively modeling these factors is essential for achieving robust and reliable 3D object detection across diverse weather conditions and distances.

\section{Methodology}
\label{sec:method}

\subsection{Problem Statement and Overall Architecture}
\label{ssec:methor_overview}
ATN3D is a LiDAR-Radar-based early 3D object detection method for an autonomous system. The overall architecture of the ATN3D is illustrated in Figure \ref{fig:atn3d_arch}. The LiDAR point clouds are denoted with $\mathcal{P}^l=\{p_i^{l}\}_{i=1}^{N_l},$ and the Radar point clouds are with $\mathcal{P}^r=\{p_i^{r}\}_{i=1}^{N_r}$. Here, $N^l$ and $N^r$ denote the number of LiDAR and Radar points in the scene. We begin by denoising the Radar input data with FDM, since the Radar input can be noisy and contain numerous background points. Hence, it is essential to reduce false-positive Radar points for improved performance. To achieve this goal, we utilize a simple semantic classification-based denoising technique \cite{huang2025l4dr} with a segmentation head ($\mathcal{S}=F_s(\mathcal{P}^r)$) to predict foreground semantic probability for each point in the 4D Radar data. The points below a predefined threshold are filtered out from further consideration. The operation can be formulated as follows: $\mathcal{P}^r_{new}= p^r_i| \mathcal{S}_i > \theta$, with $\theta$ being foreground threshold.

Next, we pass the curated Radar and LiDAR input to the density-aware early LiDAR-Radar fusion with cross-modal gating (DA-fusion) module. The fused features ($F_{feat}$) are rich with properties from both modalities for efficient feature learning. Next, the fused features are fed into the occupancy-gated neighborhood aggregation with circular BEV kernels (O-GNA) module to address the voxel sparsity issue within a proximity defined as a hyperparameter. The aggregated features are denoted with $AGG_{feat}$. Next, we introduce a lightweight evidence-conditioned channel self-attention (E-CSA) to reweight the aggregated feature channels. Finally, the LiDAR, Radar, and $AGG_{feat}$ are passed to a multi-scale gated fusion network to produce the final BEV spatial feature. The Detection Head takes the 2D BEV feature as input, uses anchors for proposal generation, and performs classification and regression tasks.

\subsection{Density-Aware Early L-R Fusion with Cross-Modal Gating (DA-fusion)}
\label{ssec:lr_fusion}
After denoising the Radar points, we pass them to the DA-fusion module along with the LiDAR points. Here, we perform early fusion on raw sensor data. The first step is to produce pillar voxels from LiDAR and Radar input points. We choose to use pillar voxel instead of true-voxel as it is more suitable for Radar data with sensitive height information. Next, each LiDAR point comes with $\mathcal{P}^l=[\mathcal{X},\lambda]$ and 4D Radar point comes with $\mathcal{P}^r=[\mathcal{X}, \mathcal{V},\Omega, t]$ properties. Here, $\mathcal{X}=[x,y,z]$ and $\lambda$ is the LiDAR point intensity, $\mathcal{V}=[v_r, v_a]$ are respectively the relative and absolute radial velocity of Doppler information, and $\Omega$ is the Radar Cross-Section (RCS). The early fusion is performed on the voxel across modalities, where there is an overlap between the LiDAR and Radar voxels. We calculate the LiDAR-Radar voxel overlap using Euclidean distance ($\mathcal{D}[v_l, v_r]=0$). Now, to perform density-aware fusion, we extend the LiDAR and Radar voxel properties with Density and Sparsity information as follows: 

\[
\begin{aligned}
\textbf{Point counts:}\quad
\tilde n^{(m)}_i &= \operatorname{clip}\!\big(n^{(m)}_i,\,1,\,T^{(m)}\big) \\[4pt]
\textbf{Density (log norm.):}\quad
d^{(m)}_i &= \frac{\log\!\bigl(1+\tilde n^{(m)}_i\bigr)}{\log\!\bigl(1+T^{(m)}\bigr)}. \\[4pt]
\textbf{Sparsity:}\quad
s^{(m)}_i &= 1 - d^{(m)}_i. \\[6pt]
\end{aligned}
\]

Here, m $\in$ \{LiDAR, Radar\}, voxel point counts: $n^{(m)}$, and voxel cap: $T^{(m)} \in \mathbb{N}$. Following previous works \cite{Wang_2023_CVPR, huang2025l4dr}, we also add a few more important voxel geometric properties, such as distance to the arithmetic mean of all points inside the pillar voxel. We denote them by: $\mathcal{Y}^l_{cl}$ and $\mathcal{Y}^r_{cr}$. Next, the offset from the pillar center in x and y coordinates is added for both modalities. We denote these with $\mathcal{C}^l$ and $\mathcal{C}^r$. After the feature extension, the complete LiDAR and Radar voxel features become:

\begin{equation}
f^{l}_{(i,j)} = \big[\, \mathcal{X}^{l},\; \mathcal{Y}^{l}_{\mathrm{cl}},\; \mathcal{D}^l,\; \mathcal{S}^l,\; \mathcal{C}^{l},\; \lambda \,\big],
\end{equation}
and, 
\begin{equation}
f^{r}_{(i,k)} = \big[\, \mathcal{X}^{r},\; \mathcal{Y}^{r}_{\mathrm{cr}},\; \mathcal{D}^r,\; \mathcal{S}^r,\; \mathcal{C}^{r},\; \mathcal{V},\; \Omega \,\big],
\end{equation}

 \begin{figure}[!ht]
\centering
    \includegraphics[width=.47\textwidth]{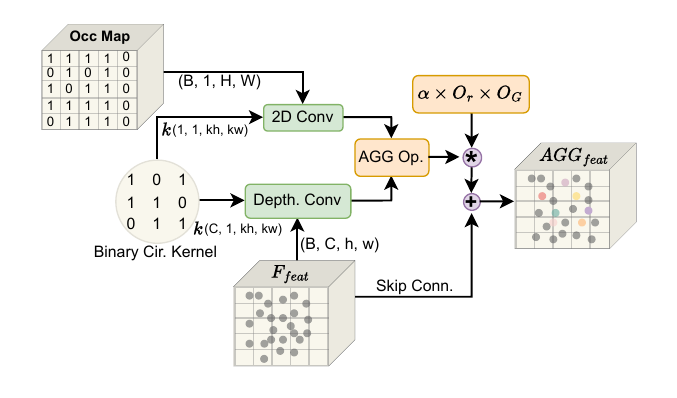}
    \caption{Occupancy-Gated Neighborhood Aggregation (O-GNA) with Voxel-Proximity Aware Circular Kernel.} 
    \label{fig:gna}
\end{figure}

Finally, bi-directional fusion happens on the extended feature by fusing $f^{l}_{(i,j)}$ and $f^{r}_{(i,k)}$ as follows: 

\begin{equation}
\begin{aligned}
\hat{f}^{\,l}_{(i,j)} &= \big[\, \mathcal{X}^{l},\; \mathcal{Y}^{l}_{\mathrm{cl}},\; \mathcal{D}^l,\; \mathcal{S}^l,\;\mathcal{C}^{l},\; \lambda,\; \overline{\mathcal{D}^r},\; \overline{\mathcal{V}},\; \overline{\Omega} \,\big],\\
\hat{f}^{\,r}_{(i,k)} &= \big[\, \mathcal{X}^{r},\; \mathcal{Y}^{r}_{\mathrm{cr}},\; \mathcal{D}^r,\; \mathcal{S}^r,\; \mathcal{C}^{r},\; \bar{\lambda},\; \overline{\mathcal{D}^l},\; \mathcal{V},\; \Omega \,\big].
\end{aligned}
\end{equation}

Here, the fusion goal is to share the same type of properties between the modalities for cross-modal fusion. The Overline denotes the average from the other modalities within the same voxel pillar. This fusion approach is beneficial because $\lambda$ and $\Omega$ are crucial for accurate classification, while velocity information, such as $v_r$ and $v_a$, is useful for differentiating dynamic objects. This bi-directional fusion of extended features makes comprehensive use of both modalities, robust to various weather and moving conditions. Later, multiple linear and max pooling layer is applied on the fused feature to obtain the corresponding BEV feature $\mathcal{F}$.

\subsection{Occupancy-Gated Neighborhood Aggregation with circular BEV Kernels (O-GNA)}
\label{ssec:gna}
The LiDAR and Radar BEV features are denoted with $\mathcal{F}_l$ and $\mathcal{F}_r$, respectively. The fused feature ($\mathcal{F}_{feat}$) is formulated by stacking both modality features in the channel dimension. The detailed architecture of the O-GNA module is illustrated in Figure \ref{fig:gna}.  We perform the neighborhood aggregation on the fused feature. O-GNA module requires two inputs: 1) the fused feature ($\mathcal{F}_{feat}$) and 2) the Voxel Occupancy Map ($Occ_{M}$). The $Occ_{M}$ is a binary map that indicates if a cell has a voxel present or not. Next, we design an adaptive circular kernel based on the defined kernel radius $k_{r}$. For the aggregation, first, we perform channel-wise convolution with the circular Kernel and the $Occ_{M}$. The goal is to sum over the neighboring voxel within the kernel radius ($k_{r}$) to gather more context information. The context summation ($num_{b,c}$) is then divided by the number of occupied voxels near the candidate voxel ($den_{b,1}$). The use of $Occ_{M}$ is very crucial; it is used to perform the gating operation during the neighborhood aggregation. We only update cells where we already have voxel information. Otherwise, the aggregation from the object boundary voxels will activate new voxels and incur noise in the feature map. Also, we do not treat all neighboring voxels equally. Voxels near the candidate voxel receive more weights than voxels at a greater distance. The complete operations are as follows:

\begin{figure}[!ht]
\centering
    \includegraphics[width=.45\textwidth]{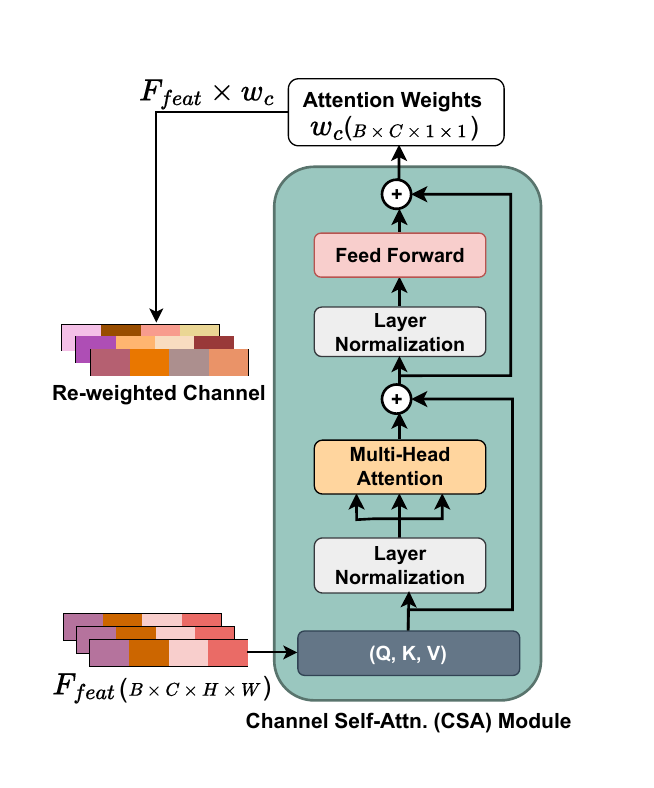}
    \caption{Evidence-Conditioned Channel Self-Attention Module for channel weight learning.} 
    \label{fig:csa}
\end{figure}

Let, Fused feat. $f_{feat}\in\mathbb{R}^{B\times C\times H\times W}$ and Cir. kernel $K\in\{0,1\}^{k_h\times k_w}$. Given an occupancy map $\mathrm{occ}\in\{0,1\}^{B\times 1\times H\times W}$, the O-GNA output $AGG_{feat}$ is computed as:
\begin{equation}
\mathrm{num}_{b,c} = K * x_{b,c}, \text{ and }\mathrm{den}_{b,1} = K * \mathrm{occ}_{b,1}
\end{equation}

\begin{equation}
\text{(norm. agg.)}\quad
\mathrm{agg}_{b,c} = \frac{\mathrm{num}_{b,c}}{\max(\mathrm{den}_{b,1},\,\varepsilon)},
\end{equation}

\begin{equation}
\rho = \Big(\tfrac{\mathrm{den}_{b,1}}{N_{\max}}\Big) \wedge 1,\text{ and }\alpha_{\mathrm{eff}}= \alpha \,\rho, \text{ and } g=\mathbf{1}\{\rho\ge \eta\}
\end{equation}

\begin{equation}
AGG_{feat} = x_{b,c} \;+\; \alpha_{\mathrm{eff},b,1}\odot g_{b,1}\odot\mathrm{agg}_{b,c}
\end{equation}

where $*$ denotes 2D convolution, $\odot$ is elementwise multiplication, $\alpha>0$ is a base gain, $\eta\in[0,1]$ is a minimum neighbor ratio, and $\varepsilon>0$ avoids division by zero.

\subsection{Evidence-Conditioned Channel Self-Attention (E-CSA)}
\label{ssec:csa}
Another novel integration in our ATN3D pipeline is performing evidence-conditioned channel self-attention on the late fusion output. The late fusion (MSGF) is utilized from the baseline model \cite{huang2025l4dr} with stride and scale modification. The MSGF performs convolution in Inter-Modal and Intra-Modal fashion. The operation is as follows:

\begin{equation}
\mathcal{F}^m_D = ReLU\big(k*\big(\mathcal{F}^m_{D-1}\big)\big)
\end{equation}

Here, $\mathcal{D} \in [1,3]$ denotes the layer number, and $m \in [l,r,agg_{feat}]$ represent different modalities. Next, a Gate operation ($\mathcal{G}$) is performed on each convolution output to reduce the possible data redundancy. The Gate Network ($\mathcal{G}$) learns adaptive gating weights $\mathcal{W}^l$, and $\mathcal{W}^r$.

\begin{equation}
\mathcal{F}_{D}^{m} = \mathcal{G}_{D}\!\left(\mathcal{F}_{D}^{m}, \mathcal{F}_{D}^{agg}\right),\; m \in \{l,r\}
\end{equation}

The output of the MSGF is a single spatial feature ($\mathcal{F}_f$) with a shape of $\mathcal{F}_f\in\mathbb{R}^{B\times C\times H\times W}$. In our design, the number of channels (C) is 768. Although we performed redundancy filtering in the MSGF module, it does not filter out or control the noise from LiDAR background points or the 4D Radar data. Hence, it is necessary to re-weight the channels before passing them into the Detection Head. Our proposed E-CSA module is illustrated in Figure \ref{fig:csa}, where we use the $\mathcal{F}_f$ from MSGF as the input. For the self-attention mechanism, we utilize the same input as Query (Q), Key (K), and Value (V). The complete E-CSA operation is as follows:

\begin{equation}
Q = \mathcal{F}_f W_Q,\quad K = \mathcal{F}_f W_K,\quad V = \mathcal{F}_f W_V
\end{equation}

\begin{equation}
\label{eq:mha}
\begin{split}
\text{Attn. per head: }\;
A_i &= \mathrm{softmax}\!\left(\frac{Q_i K_i^{\top}}{\sqrt{d}}\right) \in \mathbb{R}^{B\times C\times C}\\
Z &= \mathrm{MHA}(Q,K,V),\\
y &= \mathcal{F}_f + Z,\\
O &= y + \mathrm{FFN}\!\big(\mathrm{LN}(y)\big).
\end{split}
\end{equation}

\begin{equation}
\begin{split}
\text{Channel Weights: }\;
w_c&=\frac{C\,\tilde{w}}{\sum_{c=1}^{C}\tilde{w}_c}, \tilde{w}=\sigma(O), \qquad\\
\tilde{w}_c &= \mathrm{reshape}(w_c) \in \mathbb{R}^{B\times C\times 1\times 1},\\
F_{\text{out}} &= \mathcal{F}_{\text{f}} \odot \tilde{w}_c \in \mathbb{R}^{B\times C\times H\times W}.
\end{split}
\end{equation}

Here, $A_i$ in Eq. \ref{eq:mha} is attention in a single head. The same operation occurs in Multi-Head Attention (MHA) by splitting the input over different heads. We independently apply the Sigmoid ($\sigma$) function over the channel attention, rather than using global SoftMax, to prevent vanishing gradients caused by the weight distribution across 768 channels. The $\odot$ denotes element-wise multiplication with $\tilde{w}_c$ and the $\mathcal{F}_f$. 

\subsection{Range-Aware Loss Calculation for Distance-Stratified 3D Detection (RALC)}
\label{ssec:ralc}
Our detection head uses two sets of anchors for each class to regress and classify region proposals. As discussed in the introduction and illustrated in Figure \ref{fig:sparsity_plot}, the number of points tends to decrease significantly with range. This increases the False Negative cases and missing objects due to a low confidence score on the predictions. As one of our major goals is early and reliable detection of far-range objects, we penalize more for the incorrect classification of distant objects. To achieve this, we propose a range-aware loss calculation method that augments losses with range as a weight. We calculate the distance between the ego vehicle and anchor predictions in the XY plane as follows:

\begin{equation}
\begin{split}
\textbf{BEV radius and norm. : }
r_{b,m}=\|\mathbf{c}_m\|_2,\\
\hat r_{b,m}=\mathrm{clip}_{[0,1]}\!\left(\frac{r_{b,m}}{r_{\max}+\varepsilon}\right).
\end{split}
\end{equation}

\begin{equation}
\textbf{Range mapping: }
w^{\mathrm{range}}_{b,m}=\;1+\beta\,\hat r^{\,K}.
\end{equation}

\begin{equation}
\begin{split}
\textbf{Class weights: } y^{\mathrm{pos}}_{b,m}\in\{0,1\}, \\
w_{b,m}=(1-y^{\mathrm{pos}}_{b,m}) \;+\; y^{\mathrm{pos}}_{b,m}\cdot w^{\mathrm{range}}_{b,m}.
\end{split}
\end{equation}

\begin{equation}
\begin{split}
&\textbf{Weighted cls. loss: }
\mathcal{L}_{\mathrm{cls}}
=\dfrac{1}{B}\sum_{b=1}^{B}\sum_{m=1}^{M} w_{b,m}\;\ell\!\left(p_{b,m},\,t_{b,m}\right).
\end{split}
\end{equation}

Where $c_m$ is the center of anchor m, and $r_{\max}$ is the max BEV radius ($\epsilon>0$ to avoid divide-by-zero). K and $\beta$ are the Range Exponent (curvature) and range penalty control. The regression loss is similarly augmented with anchor range weights. The overall loss is calculated using the following formula:

\begin{equation}
\mathcal{L}
= \beta_{\mathrm{cls}}\mathcal{L}_{\mathrm{cls}}
+ \beta_{\mathrm{loc}}\mathcal{L}_{\mathrm{loc}}
+ \beta_{\mathrm{fdm}}\mathcal{L}_{\mathrm{fdm}}.
\end{equation}

Here, $\beta_{\mathrm{loc}}$ is the regression loss and $\beta_{\mathrm{fdm}}$ is the foreground semantic classification loss. The $\beta_{cls}$, $\beta_{loc}$, and $\beta_{fdm}$ are respectively weights for classification, regression, and FDM module.

\section{Experiments}
\label{sec:exp}
In this section, we will present our experimental setup and performance evaluation for our proposed ATN3D model. We conduct a comprehensive experiment on the large-scale VoD dataset for autonomous vehicle 3D object detection. For our experiment, we focus on three classes: 1) Car, 2) Pedestrian, and 3) Cyclist from the VoD dataset. We perform experiments under various weather settings, such as: 1) Clear Weather and 2) Heavy Fog Weather. With this, we prove the robustness of the ATN3D model under various weather conditions. Next, we also perform rigorous evaluation for different radius ranges from the ego vehicle. Broadly based on the point distribution from Figure \ref{fig:gt_radius_plot} and \ref{fig:sparsity_plot}, we divide the test dataset into short ($\leq30$m) and long-range ($>30$m). The goal is to verify the early detection performance when objects are at a long distance with very sparse points compared to near objects with dense points in the voxels. Our ATN3D method is compared with several of the latest SOTA methods for 3D object detection and stands out as the highest performer in various evaluation settings. 

 \begin{figure*}[!ht]
\centering
    \includegraphics[width=.95\textwidth]{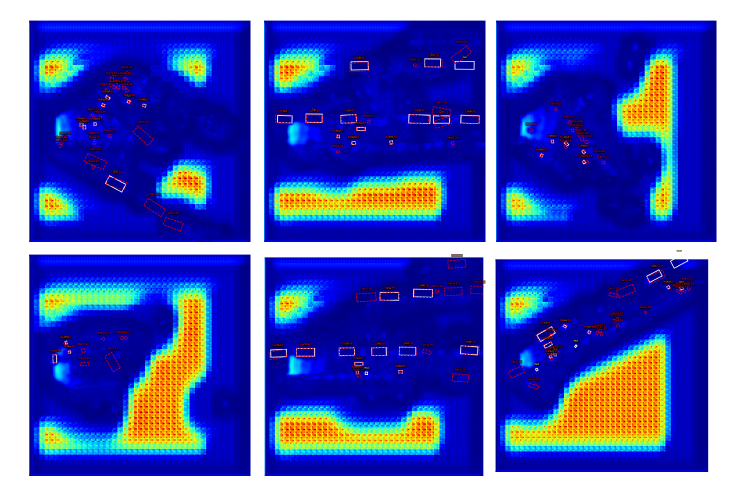}
    \caption{ATN3D detection result for a few samples on the final dense feature. Here, the white boxes represent GT, and the red boxes are model predictions with the confidence score.} 
    \label{fig:detection_plot}
\end{figure*}

\subsection{Datasets}
\label{subsec:dataset}
Our experimental VoD dataset \cite{apalffy2022} contains 8,693 64-line LiDAR, 4D Radar, and Stereo Camera frames. The official training and test dataset contains, respectively, 5139 and 1296 frames from each modality. Following the recent works, we also use the 5-scans version of the Radar data for all experiments. To explore the performance under heavy fog conditions, we use the official simulated LiDAR fog dataset with fog density $\alpha=0.03$ for mild fog and $\alpha=0.06$ for heavy fog, and kept the 4D Radar data unchanged to match the SOTA evaluation. The dataset contains more than 26,000 pedestrian, 10,000 cyclist, and 26,000 car labels. The LiDAR data includes four important information about each point: [x, y, z, intensity], whereas the 4D Radar includes: [x, y, z, azimuth, Doppler info]. The ground truth (GT) defines each object with 6 DoF $(x, y, z, w, h, \theta)$, along with additional information such as occlusion, difficulty, activity, etc. 

\subsection{Implementation Details}
\label{subsec:setup}
Our implementation is based on OpenPCDet \cite{openpcdet2020} and L4DR \cite{huang2025l4dr} code base. We use PyTorch as the Python Neural Network framework. Following the baseline model, we also use the common PointPillars \cite{lang2019pointpillars} base architecture for LiDAR and 4D Radar-based 3D object detection. Throughout all experiments, we keep batch-size=12, learning-rate=0.003, Training-epoch=50, and optimizer=adam\_onecycle. There are a few important hyperparameters, such as SCORE\_THRESH=0.15, NMS\_THRESH: 0.30, point cloud range= [0, -32.0, -3, 64, 32.0, 2], and Max points per voxel=32. For the O-GNA module, radius range is 0.32, $\alpha=0.7$, and the minimum neighborhood ratio is set to 0.4 as the gate threshold. Finally, the loss function weights $\beta_{cls}$, $\beta_{loc}$ and $\beta_{fdm}$ are respectively 1, 2 and 0.5. We use two different evaluation metrics on the VoD dataset. One is the KITTI official metrics, which are used to analyze performance under different thresholds and difficulties. We also use the official performance metrics provided with the VoD dataset, focusing on average performance to compare the SOTA methods. We conduct all our experiments on an RTX A6000 GPU.

\subsection{Performance Measures and Comparisons}
\label{subsec:measures}
In this section, we compare our proposed ATN3D method with several state-of-the-art 3D object detectors for autonomous vehicles. To present the influence of different modality inputs, we include methods that use single-modal LiDAR-only detection as well as multimodal LiDAR+Radar detection. The included SOTA methods are briefly introduced below:

\begin{table*}[t]
\centering
\setlength{\tabcolsep}{4pt}
\renewcommand{\arraystretch}{1.15}
\caption{Quantitative results of different methods on VoD-Fog dataset with the \textsc{KITTI} metric under various fog levels.}
\label{tab:vod_sota_results}
\resizebox{0.95\textwidth}{!}{%
\begin{tabular}{@{}c l c
*{9}{r}@{}}
\toprule
\multirow{2}{*}{\makecell{Fog\\Level}} &
\multirow{2}{*}{Methods} &
\multirow{2}{*}{Modality} &
\multicolumn{3}{c}{Car (IoU=0.5)} &
\multicolumn{3}{c}{Ped. (IoU=0.25)} &
\multicolumn{3}{c}{Cyclist (IoU=0.25)} \\
\cmidrule(lr){4-6}\cmidrule(lr){7-9}\cmidrule(l){10-12}
& & & \makecell{Easy} & \makecell{Mod.} & \makecell{Hard}
  & \makecell{Easy} & \makecell{Mod.} & \makecell{Hard}
  & \makecell{Easy} & \makecell{Mod.} & \makecell{Hard} \\
\midrule

\multirow{6}{*}{\makecell{W/o Fog}} 
& PointPillars & L      & 84.9 & 73.5 & 67.5 & 62.7 & 58.4 & 53.4 & 85.5 & 79.0 & 72.7 \\
& SAFNet       & L      & 85.7 & 77.2 & 69.6 & 65.5 & 64.0 & 60.2 & 90.2 & 84.6 & 83.4 \\
& InterFusion  & L+4DR  & 67.6 & 65.8 & 58.8 & 73.7 & 70.1 & 63.7 & 90.3 & 86.0 & 81.2 \\
& LiRaFusion   & L+4DR  & 84.4 & 76.0 & 67.3 & 70.2 & 69.6 & 64.9 & 88.8 & 87.2 & 82.8 \\
& L4DR         & L+4DR  & 85.0 & 76.6 & 69.4 & 74.4 & 72.3 & 65.7 & 93.4 & 90.4 & 83.0 \\
& ATN3D (Ours) & L+4DR  & \textbf{87.5} & \textbf{79.2} & \textbf{72.0} & \textbf{77.6} & \textbf{75.5} & \textbf{66.9} & \textbf{95.7} & \textbf{93.4} & \textbf{84.8} \\
\midrule

\multirow{6}{*}{Mild Fog} 
& PointPillars & L      & 79.9 & 72.7 & 67.0 & 59.9 & 55.6 & 50.5 & 85.5 & 78.2 & 72.0 \\
& SAFNet       & L      & \textbf{80.6} & 74.1 & 68.5 & 69.8 & 64.0 & 61.5 & 87.2 & 81.4 & 77.6 \\
& InterFusion  & L+4DR  & 66.1 & 64.0 & 56.9 & 74.0 & 70.6 & 64.5 & 91.6 & 87.4 & 82.0 \\
& LiRaFusion   & L+4DR  & 78.3 & 72.5 & 67.0 & 74.5 & 70.8 & 64.8 & 91.9 & 89.5 & 80.4 \\
& L4DR         & L+4DR  & 77.9 & 73.2 & 67.8 & 75.4 & 72.1 & 66.7 & \textbf{93.8} & \textbf{91.0} & 83.2 \\
& ATN3D (Ours) & L+4DR  & 80.1 & \textbf{77.8} & \textbf{70.6} & \textbf{77.0} & \textbf{74.6} & \textbf{68.2} & 93.7 & 90.8 & \textbf{83.9} \\
\midrule

\multirow{6}{*}{Heavy Fog} 
& PointPillars & L      & 67.0 & 51.4 & 44.4 & 53.1 & 47.2 & 42.7 & 69.6 & 62.7 & 57.2 \\
& SAFNet       & L      & 67.6 & 54.0 & 46.5 & 60.4 & 55.0 & 53.6 & 78.8 & 74.3 & 69.0 \\
& InterFusion  & L+4DR  & 56.0 & 48.5 & 41.5 & 63.2 & 57.8 & 52.9 & 77.3 & 71.1 & 66.2 \\
& LiRaFusion   & L+4DR  & 66.5 & 55.2 & 47.5 & 62.7 & 58.0 & 54.2 & 81.8 & 70.0 & 69.6 \\
& L4DR         & L+4DR  & 68.5 & 56.4 & 49.3 & 63.1 & 59.9 & 55.1 & 82.7 & 70.8 & 70.7 \\
& ATN3D (Ours) & L+4DR  & \textbf{72.3} & \textbf{63.0} & \textbf{54.4} & \textbf{67.5} & \textbf{63.0} & \textbf{58.7} & \textbf{85.8} & \textbf{75.2} & \textbf{73.3} \\
\bottomrule
\end{tabular}%
}
\end{table*}

\noindent\textbf{PointPillars \cite{lang2019pointpillars}} is the first work to propose a new representation for point cloud data. This work utilizes PointNets to learn a representation of point clouds organized in vertical columns (pillars). The encoded features can be used with any standard 2D convolutional detection architecture. It shows significant performance improvements over fixed encoders.

\noindent\textbf{SAFDNet \cite{zhang2024safdnet}} Existing SOTA 3D object detectors rely on dense features in the backbone network and prediction head, which gradually increases the computational cost as the perception range increases, making these models hard to scale up for long-range detection. SAFDNet proposes a fully-sparse LiDAR-only 3D detection network for computational efficiency and faster detection.

\noindent\textbf{InterFusion \cite{wang2022interfusion}} is a multimodal LiDAR+Radar based 3D perception model. It argues that the existing methods cannot learn interactions from different modalities and do not utilize the modalities properly. Utilizing the self-attention mechanism, the author proposes an interaction-based fusion framework to fuse 16-line LiDAR with 4D Radar. It aggregates features from two modalities and identifies cross-modal relations between Radar and LiDAR features.

\noindent\textbf{LiRaFusion \cite{song2024lirafusion}} is one of the latest LiDAR+Radar-based framework that efficiently perform cross-modal fusion across the network. For early fusion, it stacks the LiDAR and Radar point clouds and learns feature representation through a joint encoder. For the late fusion, it uses an adaptive gated network to learn modality interaction and control data redundancy. 

\noindent\textbf{L4DR \cite{huang2025l4dr}} is our baseline model. We choose this model as the baseline due to its superiority in joint feature learning. For early fusion, it proposes to use both modalities separately and perform voxel-level fusion. In the late fusion, it proposes Inter-fusion as well as Intra-fusion for the fullest use of each sensor. The gate operation helps to reduce redundant data due to the Inter-Intra fusion operation. 

\textbf{SOTA performance comparison for VoD Dataset:} The SOTA performance comparison for the VoD dataset with KITTI metrics is presented in Table \ref{tab:vod_sota_results}. For a comprehensive performance comparison, we use two LiDAR-only methods and four L+4DR methods. The goal is to present the performance gain with 4D Radar in adverse weather conditions. The KITTI metrics divide the performance into three difficulties for each class. The post-processing IOU for car, pedestrian, and cyclist are 0.5, 0.25, and 0.25, respectively.

We start the comparison with the LiDAR-only PointPillars model, which shows decent performance for the Car and Cyclist classes. The 3D mAP for cars and cyclists on average (diff.) is 74.66\% and 79.06\%, whereas for pedestrians the mAP is 58.16\% in W/o fog condition. The performance drop comes from the inability to perform rich feature extraction for the small-sized objects. For adverse weather conditions, the performance drops gradually with increasing fog intensities. The SAFNet is a recent LiDAR-only model that shows competitive performance for clear and mild weather compared to even L+4DR models. It achieves 77.5\% and 86.0\% average difficulty mAP, respectively, for car and cyclist, matching the baseline L4DR performance. It shows the effectiveness of its adaptive feature diffusion strategy for an improved sparse network. However, the performance in heavy fog conditions and with small objects was not satisfactory. The performance decreases by nearly 20\% across classes compared to the clear weather performance. The performance of LiRAFusion and L4DR is very close in clear weather and mild fog settings. We notice a slightly better performance under mild fog conditions for the pedestrian class. The performance gain for the pedestrian class with L4DR compared to single modal methods is approximately $+\Delta9\%$ under all weather conditions. This shows the superiority of Radar data when detecting small objects under adverse weather conditions. 

\begin{table}[t]
\centering
\scriptsize
\setlength{\tabcolsep}{5pt}
\renewcommand{\arraystretch}{1.15}
\caption{Performance comparison on Short vs. Long Distance Range with VoD 3D mAP metric in Clear Weather.}
\label{tab:perm_comp_clear}
\begin{tabular}{@{}l c | r r r | r r r@{}}
\toprule
\multirow{2}{*}{Methods} & \multirow{2}{*}{Modality}
& \multicolumn{3}{c|}{Short Dist. ($\leq$30m)} & \multicolumn{3}{c}{Long Dist. ($>$30m)} \\
& & Car & Ped. & Cyc. & Car & Ped. & Cyc. \\
\midrule
Pointpillars         & L        & 69.25 & 53.56 & 71.10 & 58.69 & 50.65 & 60.02 \\
SAFNet               & L        & 72.30 & 58.60 & 80.66 & 60.10 & 52.22 & 61.48 \\
InterFusion          & L+4DR    & 58.26 & 64.42 & 86.36 & 51.42 & 58.32 & 67.64 \\
LiRaFusion           & L+4DR    & 70.82 & 61.32 & 80.28 & 62.46 & 56.70 & 64.40 \\
L4DR                 & L+4DR    & 71.88 & 69.50 & 88.21 & 64.62 & 59.47 & 70.92 \\
\textbf{ATN3D (Ours)} & L+4DR   & \textbf{73.14} & \textbf{72.08} & \textbf{90.30} & \textbf{68.12} & \textbf{64.18} & \textbf{72.69} \\
\bottomrule
\end{tabular}
\end{table}

However, the best performance comes from the proposed ATN3D method (Table \ref{tab:vod_sota_results}). For Car--Easy, it not only achieves the top clear score (87.5) but also preserves more performance in heavy fog (72.3, a -15.2 drop) than any baseline (e.g., L4DR -16.5, LiRaFusion -17.9, LiDAR-only $\approx$-18). The advantage is most visible for Pedestrian and Cyclist, where targets are smaller, often partially occluded, and where LiDAR degradation most harms localization: ATN3D’s Cyclist–Easy score in heavy fog reaches 85.8, clearly above the best baseline (82.7, L4DR). For the challenging pedestrian class, the ATN3D achieves mAP of 63\% on avg. for heavy fog conditions, beating the baseline by 4\%. Our ATN3D outperforms the baseline L4DR by utilizing more context information with the O-GNA module and the baseline's rich late fusion network. The qualitative performance is illustrated in Figure \ref{fig:detection_plot}, where we visualize the final dense feature with GT boxes (white) overlayed with predictions (red). In Figure \ref{fig:detection_plot}, it is evident that our model successfully detects most of the objects in the scene, missing only a few very small pedestrian objects. Our experiments found that many missed objects consist of only a single voxel, making the O-GNA module inapplicable.

\begin{table}[t]
\centering
\scriptsize
\setlength{\tabcolsep}{5pt}
\renewcommand{\arraystretch}{1.15}
\caption{Performance comparison on Short vs. Long Distance Range with VoD 3D mAP metric in Heavy Fog Weather.}
\label{tab:perm_comp_heavy_fog}
\begin{tabular}{@{}l c | r r r | r r r@{}}
\toprule
\multirow{2}{*}{Methods} & \multirow{2}{*}{Modality}
& \multicolumn{3}{c|}{Short Dist. ($\leq$30m)} & \multicolumn{3}{c}{Long Dist. ($>$30m)} \\
& & Car & Ped. & Cyc. & Car & Ped. & Cyc. \\
\midrule
Pointpillars    & L            & 32.33 & 42.65 & 53.14 & 18.79 & 12.30 & 16.50 \\
SAFNet          & L            & 34.40 & 44.32 & 58.25 & 19.54 & 13.36 & 17.05 \\
InterFusion     & L+4DR        & 28.06 & 46.45 & 59.63 & 20.31 & 14.80 & 17.96 \\
LiRaFusion      & L+4DR        & 33.50 & 45.67 & 61.46 & 21.30 & 14.56 & 17.76 \\
L4DR            & L+4DR        & 34.76 & 47.61 & 60.88 & 22.97 & 14.65 & 19.71 \\
\textbf{ATN3D (Ours)} & L+4DR  & \textbf{36.54} & \textbf{49.63} & \textbf{61.48} & \textbf{24.08} & \textbf{16.78} & \textbf{22.74} \\
\bottomrule
\end{tabular}
\end{table}

\textbf{Distance-based SOTA performance comparison:} One of the main focuses of our work is to improve performance for objects at far distances. Detecting far objects at an early stage is one of the major safety factors for autonomous vehicles, as the vehicle's trajectory and smooth motion planning depend on intercepting objects as early as possible. Although, as discussed, detecting objects at a far distance is often challenging, as the objects might look smaller in shape. To resolve this issue, our ATN3D took several safety guards, such as O-GNA, RALC, and in this subsection, we are going to evaluate the ATN3D using distance-filtered VoD 3D mAP metrics. We separate the test-set GT and the model predictions based on short distance ($\leq30m$) and long distance ($>30m$) Radius. Then, we pass the filtered GT and Predictions for corresponding short/long distance evaluation using the official VoD 3D mAP metrics.

\begin{table}[t]
\centering
\scriptsize
\setlength{\tabcolsep}{4pt}
\renewcommand{\arraystretch}{1.15}
\caption{Effect of each component proposed in ATN3D on VoD-Fog dataset (3D mAP + Mod. difficulty).}
\label{tab:ablation_components}
\begin{tabular}{@{}cccc | ccc@{}}
\toprule
\multicolumn{4}{c|}{\textbf{Module}} 
& \multicolumn{3}{c}{\textbf{3D mAP}} \\
\makecell{DA-Fusion} & \makecell{O-GNA} & \makecell{E-CSA} & \makecell{RALC}
& \makecell{W/o\\Fog} & Mild Fog & Heavy Fog \\
\midrule
           &  \textbf{L4DR}  &            &       & 72.54 & 68.59 & 61.26 \\
\checkmark &            &            &            & 73.32 & 70.12 & 63.35 \\
\checkmark & \checkmark &            &            & 74.88 & 71.22 & 66.92 \\
\checkmark & \checkmark & \checkmark &            & 75.65 & 71.89 & 68.15 \\
\checkmark & \checkmark & \checkmark & \checkmark & \textbf{76.09} & \textbf{74.35} & \textbf{69.67} \\
\bottomrule
\end{tabular}
\end{table}

Table \ref{tab:perm_comp_clear} focuses on distance-based performance for clear weather. From Table \ref{tab:perm_comp_clear}, we see that there is a sharp decrease in performance for every SOTA method when evaluating objects more than 30m distance. The drop is higher for the LiDAR-only model, more than 10\%. The exception is InterFusion, which shows poor performance for the car; however, it maintains good performance for the cyclist. LiRaFusion shows satisfactory results for the short distance; however, L4DR is the most consistent, leading other prior methods for both short and long ranges (e.g., Pedestrian–Short 69.50 and Pedestrian–Long 59.47). ATN3D achieves the best results in every setting, improving over the strongest baseline (L4DR) by clear margins at long range—Car (68.12 vs. 64.62, +3.5), Pedestrian (64.18 vs. 59.47, +4.7), and Cyclist (72.69 vs. 70.92, +1.8)—and also at short range, ATN3D maintains at least $+2\%$ gain across every class. These gains indicate that ATN3D not only improves near-field recognition but, more importantly, preserves accuracy at longer distances where geometry is sparse and Radar cues become faint.

Table \ref{tab:perm_comp_heavy_fog} presents the most challenging scenario, the long vs short distance performance under heavy fog conditions. The goal is to present the breaking point of the 3D detection model and push forward for improved performance under challenging conditions. Under heavy fog, LiDAR returns are strongly attenuated and scattered by droplets, causing the signal to decay with range, and near-field backscatter further reduces the signal-to-noise ratio. This leads to a sharp drop in performance for short $\rightarrow$ long distance, especially for small objects. For L4DR, the AP falls by on average 28.39\%, and for ATN3D it falls by 26\%. Despite the severity, ATN3D outperforms L4DR at both ranges, with the largest margins at long distance (+1.11/+2.13/+3.03 for Car/Ped/Cyc). The gain pivots on the O-GNA module, which performs distance-weighted voxel aggregation, and the RALC, which up-weights far-range instances during training, countering SNR collapse and long-range imbalance to boost recall.

\subsection{Ablation study}
\label{subsec:Ablation}
In this section, we answer several questions regarding the effectiveness of the proposed modules in our ATN3D framework. The first question is: Does the density-based feature extension (DA-Fusion) help to improve feature learning in early fusion? Table \ref{tab:ablation_components} shows that the ATN3D scored 73.32\% and 63.35\% of mAP under clear and heavy fog conditions, which is respectively +0.78\% and +2.09\% gain compared to the baseline L4DR model. The gain is high for heavy fog conditions where voxel density is lower and sparsity is high. The second question is: Is the O-GNA module effective for various weather conditions? In Table \ref{tab:ablation_components}, we see that the gain is very significant when using O-GNA in the pipeline. It is +1.56\% and 3.57\%, respectively, for clear weather and heavy fog conditions. In the O-GNA, we use occupancy gating and occupancy ratio weighting. 
\begin{figure}[!ht]
\centering
    \includegraphics[width=.45\textwidth]{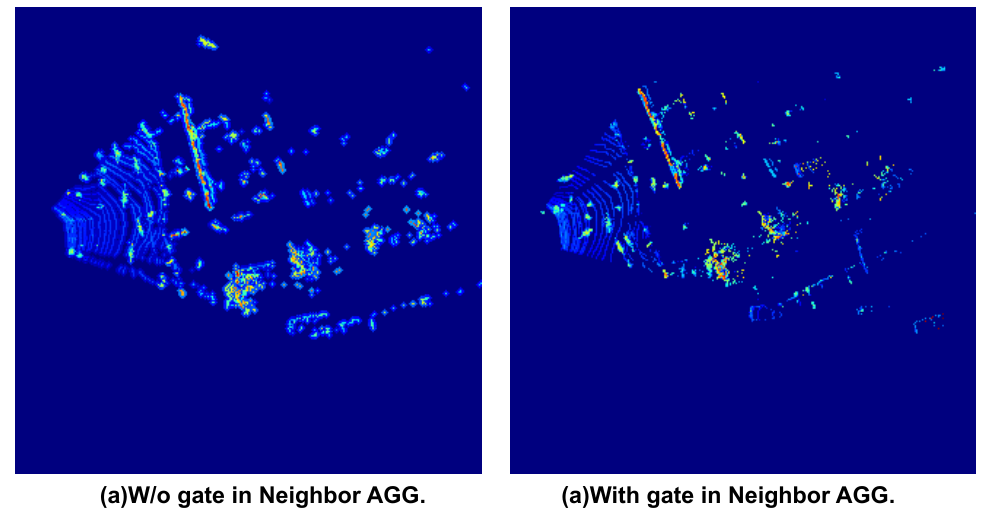}
    \caption{O-GNA module ablation on gated operation. The aggregation without a gated operation introduces noise around the boundary of the object voxels.} 
    \label{fig:gna_ablation}
\end{figure}
The reason behind the operation is illustrated in Figure \ref{fig:gna_ablation}. Without a gating operation, the aggregation operation activates empty cells and introduces noise in the feature. The effect is much visible near the edge of each object, which degrades the overall performance. Hence, the gating operation is required to keep the total activated cells the same. 

Next, to answer the question of the effectiveness of the E-CSA module, we focus on the 3rd row of the Table \ref{tab:ablation_components}. The 3rd row shows the performance of ATN3D with the DA-Fusion+O-GNA+E-CSA module. Although the performance gain for clear and mild fog weather was not significant ($<1\%$), the performance gain for heavy fog is nearly +1.23\%. The E-CSA is a self-attention network that learns channel weights for the final dense feature, which has 768 channels. Under heavy fog conditions, many channels focus on the background noise and fog droplets, which are less important for the detection task. The O-GNA is more effective in this scenario to focus more on channels relevant to the downstream task. The final question is: Is RALC effective for the ATN3D framework? We use the last of the Table \ref{tab:ablation_components} to prove the concept. Table \ref{tab:ablation_components} shows that ATN3D gains around +0.44\%, +2.46\%, and +1.52\%, respectively, for clear, mild, and heavy fog conditions. It is proven that the RALC is consistently successful under all weather conditions. However, as the RALC is specifically designed to perform well for far objects, we use Table \ref{tab:ablate_gna_ralc} to further investigate the RALC's effectiveness.

\textbf{Hyper-parameter ablation for O-GNA and RALC module:} The purpose of Table \ref{tab:ablate_gna_ralc} is to show the effect of some major hyperparameters for different weather and distance. For RALC, we mainly track $\beta$ = range penalty control and $K$ = range exponent. The baseline L4DR achieves 47.75\% and 19.11\% for short and long distances without the RALC module. Using RALC with $\beta=0.5,\text{ and } K=1$ has little effect for both cases ($\approx$+0.50\%). However, increasing curvature to $K=2$ achieves a significant gain of +2.09\% for far objects compared to the baseline model. The increase in K does not hurt the performance for the near objects. We tried penalizing more for the far objects with $\beta=0.7$; however, it shows an adverse effect on the near objects and decreases the performance by -1.18\%. Increasing the value of $K$ makes the performance worse; hence, we choose to use $\beta=0.5, \text{ and } K=2$ in our final version of ATN3D experiments.

\newcommand{\hdrbox}[1]{\makebox[7mm][c]{\(#1\)}}
\newcommand{\boxsize}[1]{\makebox[12mm][c]{\(#1\)}}

\begin{table}[t]
\centering
\scriptsize
\setlength{\tabcolsep}{4pt}
\renewcommand{\arraystretch}{1.05}
\caption{Hyper-parameter Ablation for \textbf{O-GNA} and \textbf{RALC}}
\label{tab:ablate_gna_ralc}

\begin{tabular}{@{}ccc|cc@{}}
\toprule
\multicolumn{3}{c|}{\textbf{RALC (Hyper-parameter)}} & \multicolumn{2}{c}{\textbf{VoD 3D mAP}} \\
\boxsize{\boldsymbol{\beta}} & \boxsize{\boldsymbol{K}} & \boxsize{\text{—}} & Short ($\le$30m) & Long ($>$30m) \\
\cmidrule(lr){1-5}
0.0 & 0 &  & 47.75 & 19.11 \\
0.5 & 1 &  & 48.30 & 19.56 \\
0.5 & 2 &  & \textbf{49.21} & \textbf{21.20} \\
0.7 & 1 &  & 46.57 & 19.77 \\
\midrule
\multicolumn{3}{c|}{\textbf{GNA (Hyper-parameter)}} & \multicolumn{2}{c}{\textbf{VoD 3D mAP}} \\
\hdrbox{\boldsymbol{\alpha}} & \hdrbox{\boldsymbol{\eta}} & \hdrbox{\boldsymbol{k}_{r}} & W/o fog & Heavy Fog \\
\cmidrule(lr){1-5}
--  & --  & --   & 72.54 & 61.26 \\
0.7 & 0.4 & 0.32 & 76.09 & \textbf{69.67} \\
0.7 & 0.4 & 0.64 & 74.15 & 63.45 \\
0.7 & 0.6 & 0.32 & \textbf{77.60} & 63.88 \\
0.7 & 0.6 & 0.64 & 77.48 & 60.71 \\
\bottomrule
\end{tabular}
\end{table}

For the O-GNA module, we track $\alpha, \eta \text{ and } k_r$, which are respectively gain control, occupancy ratio threshold, and kernel radius (m). We start with the baseline L4DR without the O-GNA module. It achieves 72.54\% and 61.26\% mAP for clear and heavy fog conditions, respectively. Using a circular Kernel of $k_r=0.32m$ with $\alpha=0.7 \text{ and }\eta=0.4$ achieves the SOTA performance with a gain of +3.55\% for clear and +8.41\% for heavy fog conditions. Next, we increase the kernel size to $k_r=0.64m$, keeping other parameters the same, which drops the performance. The negative effect is contributed mainly by small objects, which have limited voxel representations. Next, we increase the occupancy threshold to $\eta=0.6$. While a higher $\eta$ improves clear-weather performance (77.60\%), it causes a marked decline under heavy fog. The reason is that $\eta$ acts as a gating threshold: raising it excludes many low-occupancy voxels within the kernel radius from the context aggregation. Because heavy-fog scenes are dominated by such low-occupancy voxels, the stricter gate discards substantial useful context, leading to performance degradation. Throughout all experiments, we use $\alpha=0.7, \eta=0.4  \text{ and }k_r=0.32$, which gives stable results across various weather conditions.

\section{Conclusion and Future Work}
\label{sec:con}
This study investigated the impact of voxel-level sparsity on 3D object detection across diverse weather and range conditions. We demonstrated that sparsity becomes particularly critical under heavy fog and at long distances, where the number of LiDAR returns drops drastically, degrading voxel-level representation quality. To address these issues, we proposed ATN3D, a LiDAR–Radar 3D detection framework tailored for robust perception under extreme sparsity.

The proposed ATN3D pipeline integrates four key modules. The DA-fusion module improves early feature quality by conditioning fusion on per-voxel density and sparsity, while the O-GNA module selectively aggregates contextual information only from credible occupied voxels, effectively suppressing noise propagation. The E-CSA module adaptively reweights channels to emphasize task-relevant and weather-robust cues, and the RALC enhances long-range detection by rebalancing classification and regression objectives according to object distance. Experimental results on the VoD dataset demonstrate that ATN3D achieves consistent performance gains across both clear and adverse-weather scenarios, with up to +3.57\% mAP improvement under heavy fog and more than +2\% mAP gain for long-range targets. These findings highlight the importance of explicit sparsity modeling for reliable LiDAR–Radar early perception in intelligent transportation systems. Future work will focus on exploring point-cloud densification and object completion to further improve 3D perception robustness under challenging sensing conditions.


\ifCLASSOPTIONcaptionsoff
  \newpage
\fi

\bibliographystyle{unsrt} 
\bibliography{bib3d}

\begin{IEEEbiography}[{\includegraphics[width=1in,height=1.25in,clip,keepaspectratio]{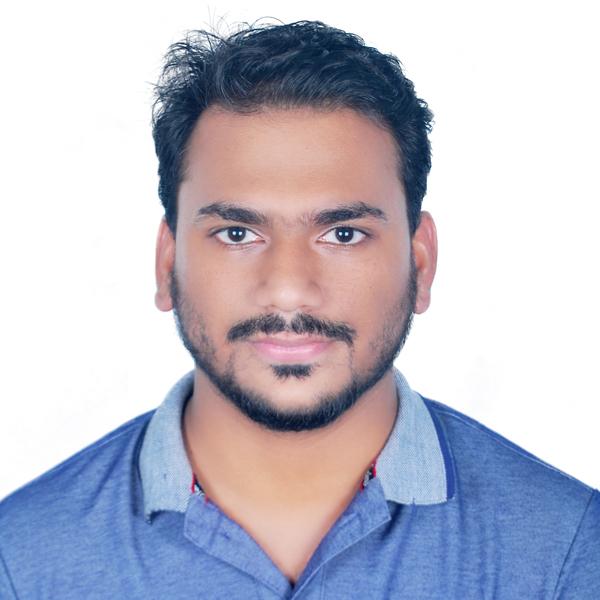}}]{Debojyoti Biswas}
received the Ph.D. degree in Computer Science from Texas State University, San Marcos, TX, USA. He is currently a postdoctoral researcher in Larson Transportation Institute at Penn State University. He received his B.Sc. in Information and Communication Engineering from Noakhali Science and Technology University, Bangladesh. He worked as a Lecturer in the Computer Science department at Leading University, Bangladesh, from 2019 to 2021, before commencing his Ph.D. studies. He is a peer reviewer for several prestigious journals, including IEEE TGRS, IEEE JSTARS, IEEE GRSL, Springer PFGE, Elsevier ESA, and others. His research interests include computer vision, video understanding, intelligent transportation systems, object detection, and domain adaptation.
\end{IEEEbiography}

\begin{IEEEbiography}[{\includegraphics[width=1in,height=1.25in,clip,keepaspectratio]{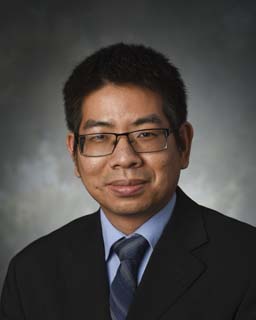}}]{Xianbiao Hu}
received the Ph.D. degree from The University of Arizona, Tucson, AZ, USA, in 2013. He is currently an Associate Professor with the Civil and Environmental Engineering Department, The Pennsylvania State University. His current research interests include smart mobility systems, connected and automated vehicles, electric vehicles, mobility behavior management, transportation big data analytics, and traffic flow and system modeling. He is an Associate Editor of IEEE Transactions on Intelligent Transportation Systems.
\end{IEEEbiography}

\end{document}